# New approach using Bayesian Network to improve content based image classification systems

Khlifia Jayech [1]   and   Mohamed Ali Mahjoub [2]

[1] SID Laboratory, National Engineering School of Sousse
Sousse, Technology Park 4054 Sahloul, Tunisia

[2] Preparatory Institute of Engineering of Monastir
Monastir, Ibn Eljazzar 5019, Tunisia

## Abstract

This paper proposes a new approach based on augmented naïve Bayes for image classification. Initially, each image is cutting in a whole of blocks. For each block, we compute a vector of descriptors. Then, we propose to carry out a classification of the vectors of descriptors to build a vector of labels for each image. Finally, we propose three variants of Bayesian Networks such as Naïve Bayesian Network (NB), Tree Augmented Naïve Bayes (TAN) and Forest Augmented Naïve Bayes (FAN) to classify the image using the vector of labels. The results showed a marked improvement over the FAN, NB and TAN.

***Keywords:*** *Bayesian Network, TAN, FAN, Image classification, Recognition, CBIR.*

## 1. Introduction

In the few last years, the domain of indexation by the content called CBIR (Content Based Image Retrieval), which includes all that works developed to search and classify images through their internal content, has been a very active domain of research. Much of the related work on image recognition and classification for indexing, classifying and retrieval has focused on the definition of low-level descriptors and the generation of metrics in the descriptor space [1]. These descriptors are extremely useful in some generic image classification tasks or when classification is based on query by example. However, if the aim is to classify the image using the descriptors of the object content this image. There are two questions to be answered in order to solve difficulties that are hampering the progress of research in this direction. First, how shall we semantically link objects in images with high-level features? That means how to learn the dependence between objects that reflect better the data? Second, how shall we classify the image using the structure of dependence finding?

Our paper presents a work which uses three variants of naïve Bayesian Networks to classify image of faces, using the structure of dependence found between objects.
This paper is divided as follows:
Section 2 presents the principal works related to CBIR and a state of art of 2D face recognition; Section 3 describes the developed approach based in Naïve Bayesian Network, we describe how the spatial features are extracted; and we introduce the method of building the Naïve Bayesian Network, Tree Augmented Naïve Bayes (TAN) and Forest Augmented Naïve Bayes (FAN), and inferring posterior probabilities out of the network; Section 4 presents some experiments; finally, Section 5 presents the discussion and conclusions.

## 2. Related works

### 2.1 State of the art in 2D Face Recognition

Face recognition is a huge research area and is the preferred mode of identity recognition by humans: it is natural, robust. Each year, the attempted solutions and algorithm grow in complexity and execution time.
Zhao et al. in [28] divide the existing algorithms into three categories, depending on the information they use to perform the classification: appearance-based methods (also called holistic), feature-based and hybrid. Sharavanan et al. in [42] present taxonomy of face recognition methods used as shown in the figure 1.

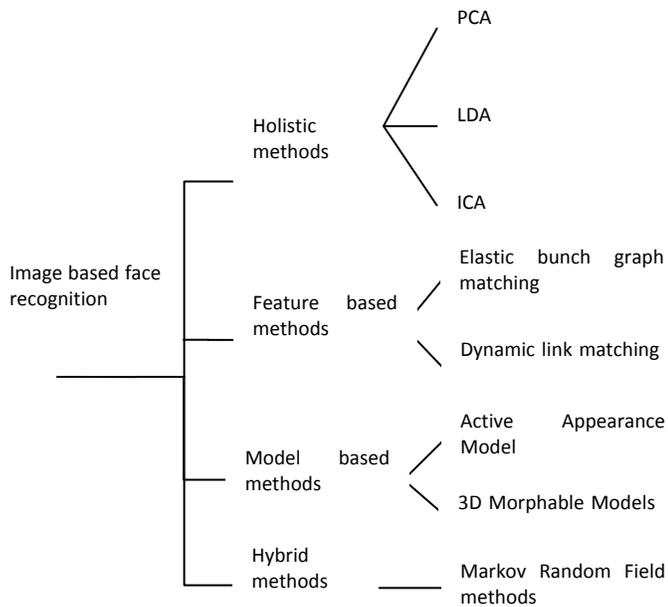

Fig. 1 Taxonomy of face recognition methods used [42].

In this section, we review the relevant work on 2D face recognition, and discuss the merits of different representations and recognition algorithms.

### 2.1.1 Holistic Approach

Holistic algorithms use global features of the complete face. Examples of holistic systems include Principal Component Analysis (PCA) [31], Linear Discriminant Analysis (LDA) [33] as well as Independent Component Analysis (ICA) [32]. These projection techniques are used to represent face images as a lower-dimensional vector, and the classification itself is actually performed by comparing these vectors according to a metric in the subspace domain.

- #### Principal Component Analysis

Principal Component Analysis (PCA) is an approach to analyze and dimensionally reduce the data, highlighting their similarities and differences. The task of facial recognition is discriminating input signals (image data) into several classes (persons).

- #### Linear Discriminate Analysis

Linear Discriminate Analysis is effective to encode discriminatory information. The idea of LDA is grouping similar classes of data where as PCA works directly on data. It tries to find directions along which the classes are best separated by taking into consideration not only the dispersion within-classes but also the dispersion between classes.

- #### Independent Component Analysis

Independent component analysis (ICA) is a generative model, expressing how observed signals are mixed from underlying independent sources. This approach transforms the observed vector into components which are statically maintained as independent. The basic restriction is that the independent components must be non-Gaussian for Independent Component Analysis to be possible.

A limitation of holistic matching is that it requires accurate normalization of the faces according to position, illumination and scale. Variations in these factors can affect the global features of the face, leading inaccurate final recognition. Moreover, global features are also sensitive to facial expressions and occlusions. Holistic methods such as neural networks [24] are more complex to implement, whereby an entire image segment can be reduced to a few key values for comparison with other stored key values with no exact measures or knowledge such as eye locations.

### 2.1.2 Feature-based Approach

Feature-based algorithms use local features [26] or regions [27] of the face for recognition. Such features may be geometric measurements, particular blocks of pixels, or local responses to a set of filters.

Feature-based algorithms try to derive an individual's face's model based on local observations. Examples of such systems include the Elastic Bunch Graph Matching (EBMG) [38], recent systems using Local Binary Patterns (LBP) [34] [37], and also statistical generative models: Gaussian Mixture Models (GMM) [35] [36].

- #### Elastic Bunch Graph Matching

This technique use the structure information of a face, which shows the fact that the images of the same subject tend to deform, scale, translate, and rotate, in the plan of image. It uses the labeled graph, edges are labeled according to the distance information and nodes are labeled with wavelet coefficients in jets. This graph can be after used to produce image graph. The model graph can be deformed, scaled, translated, and rotated during the matching process. This can make the system resistant to large variation in the images.

- #### Dynamic Link Matching

In dynamic link structure, the models are presented by layers of neurons, which are labeled by jets. Jets are wavelet elements describing grey level distribution. This approach encodes information using wavelet transformations. Dynamic link structure establishes a mapping between all-in-all linked layers, and thus reduces distortion. The method uses a winner-take-all strategy once a correct mapping is obtained choosing the right model. A typical dynamic link structure is given in the figure 2.

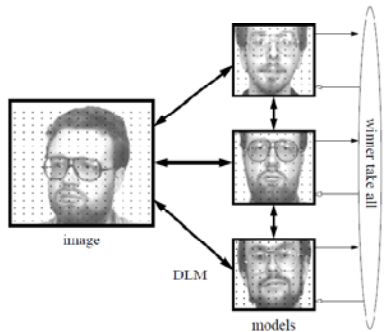

Fig. 2 Dynamic Link Architecture [42]

Face recognition systems using local features empirically show a better performance as compared to holistic algorithms [40][36]. Moreover, they also have several other advantages: first, face images are not required to be precisely aligned. This is an important property, since it increases robustness against imprecisely located faces, which is a desirable behavior in real-world scenarios.

Second, local features are also less sensitive to little variations in illumination, scale, expressions and occlusions.

To go beyond the limitations of the Eigen faces method in holistic methods, Garcia et al. in [30] present a new scheme for face recognition using wavelet packet decomposition. Each face is described by a subset of band filtered images containing wavelet coefficients. These coefficients characterize the face texture, and a set of simple statistical measures permits forming compact and meaningful feature vectors. Then, an efficient and reliable probabilistic metric derived from Bhattacharrya distance is used in order to classify the face feature vectors into person classes. The Wavelet transformation allows the generation of such facial characteristics that are invariant to lighting conditions and overlapping. It organizes the image in subgroups that are localized according to orientation and frequency. In every subgroup, each coefficient is also localized in space.

### 2.1.3 Hybrid Approach

Hybrid matching methods use a combination of global and local-features for recognition [33]. This approach is most effective and efficient in the presence of irrelevant data. The key factors affecting performance depend on the selected features and the techniques used to combine them. Feature based and holistic methods are not devoid of drawbacks. Feature based method is sometimes badly affected by accuracy problem since accurate feature localization is its very significant step. On the other hand, holistic approach uses more complex algorithms demanding longer training time and storage requirements. Huang et al in [41] proposed a hybrid face recognition method that combines holistic and feature analysis-based approaches using a Markov Random Field (MRF) model. According to the proposed method, the image is divided into smaller image patches, each having specific Ids. The model can be represented as in the figure 3. It includes two layers, observable nodes (squares in the graph, representing image patches) and hidden nodes (circles in the graph, representing the patch Ids). Edges in the graph depict relationships among the nodes.

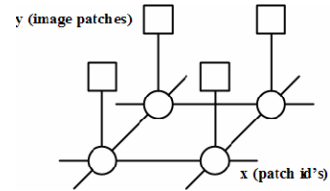

Fig. 3 MRF model [41]

### 2.2 State of the art in Bayesian classification

There are several works in CBIR systems using Bayesian Network. The majority of them involve various small problems, such as structure of Bayesian Network, feature extraction, classification and retrieval. Rodrigues in [3] presents a Bayesian network model which combines primitives' information of color, edge-map and texture to improve the precision in content-based image retrieval systems. This approach allows retrieving images according to their features. However the spatial relations between objects are not treated. Zhang in [2] suggested modeling a global knowledge network by treating and entire image as a scenario. Authors suggested a process that is divided into two stages: the initial retrieval stage and Bayesian belief network inference stage. The initial retrieval stage is focused on finding the best multi-features space, and then a simple initial retrieval is done within this stage. The Bayesian inference stage models initial beliefs on probability distributions of concepts from the initial retrieval information and construct a Bayesian belief network. In this approach, the results are largely depending on the choice of representative blocks in the first stage. However, the dependence between objects in the approach presented in [2] is also not dealt with. Mahjoub and Jayech in [4] investigated three variants of Bayesian Networks such as Naïve Bayes, Tree Augmented Naïve Bayes (TAN) and Forest Augmented Naive Bayes (FAN) in image classification. Those models study the dependence between objects. However, the images used in this study represent the structure of a document containing texts blocks and graphs.

The work presented in this paper tries to accomplish the difficult task of classifying the images of faces using the high level features. Our works study the dependence between the high level features to build an optimal structure that better represents the data saying if an object is depends or not to another by using three variants of naïve Bayesian Networks.

## 3. The proposed Approach

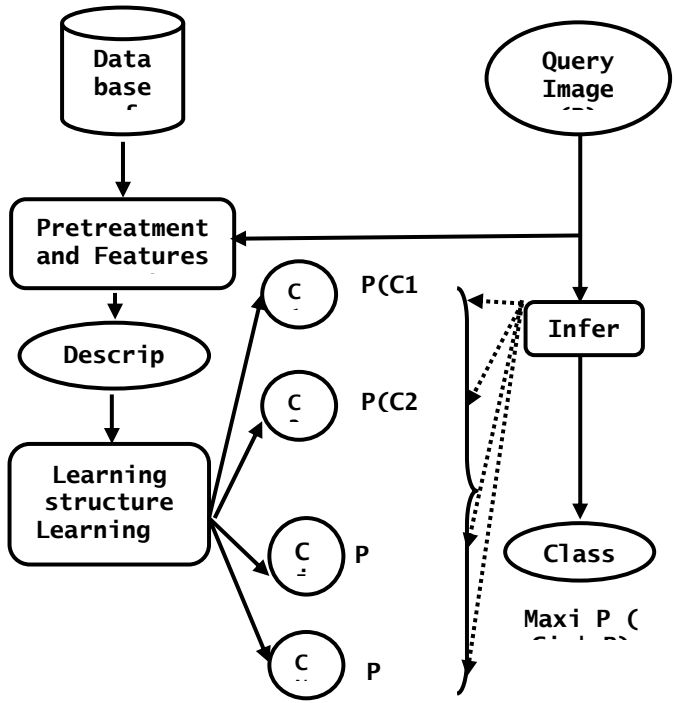

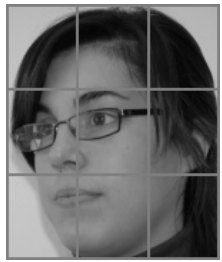

Classifying an image depends on classifying the objects within the image. The studied objects in an image are the following analysis represented by blocks. An example of one image being divided into elementary building blocks that contain different concepts is illustrated in figure 4:

Fig. 4 Example of an image divided into some elementary building blocks

For each block, we compute the histogram of color using GMM (Gaussian Mixture Model) and we compute the descriptor textural using the Grey Level Co-occurrence Matrix (GLCM). Then, we use Kmeans to cluster the object into k cluster. So each block will be labeled and integrated to one of the k cluster

### 3.1 Features Processing

- Color Feature

Color may be one of the most straight-forward features used by humans for visual recognition and discrimination. To compute this feature we used histogram of color. The histogram can be considered as a Gaussian Mixture Model [6] (figure 5) defined as:

$$f(x|\theta) = \sum_{i=1}^{K} p_i f_i(x|\alpha_i)$$

Where:

- $p_i$ is the proportion of class i ($p_i \geq 0$ and $\sum_i p_i = 1$)
- $\alpha_i = (\mu_i, \sum_i)$, $\mu_i$ and $\sum_i$ were respectively the center and the covariance of the kième normal component $f(.\,|\alpha_i)$.

The estimation of these parameters can be doing with the EM proposed by Dempster and al in 1977 and the number of k normal component can be estimated as follows:

$$BIC(K) = -2\ln\ f(X|\hat{\theta}_K) + v_K \ln n$$

The value of k is chosen between k=1 and $k_{sup}$ ($k_{sup}$ is to choose in priori) that minimizes the Bayesian criterion BIC [7]. $\hat{\theta}_K$ and $v_K$ are respectively the maximum credibility estimator and the degree of freedom of the model.

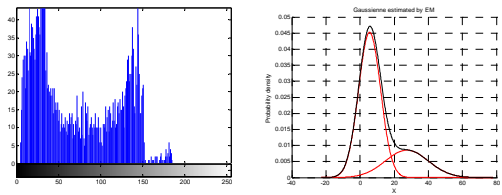

Fig. 5 Approximation of histogram by GMM

- Texture Features

Textural features are linked to the contrast of processed images. The textural features can be used to differentiate between objects having the same color. Among these features, we used the Haralick features [5]. Haralick features' computing is based on co-occurrence matrix. From these features, we use four attributes which are: energy, entropy, contrast, and homogeneity.
The energy is defined as:

$$E = \sum_{i,j} (p(i,j))^2$$

The entropy is defined as

$$ENT = -\sum_i \sum_j p(i,j) \log p(i,j)$$

The contrast is defined as

$$CONT = \sum_i \sum_j (i-j)^2 p(i,j)$$

The homogeneity is defined as

$$HOM = \sum_{i,j} \frac{1}{1+(i-j)^2} p(i,j)$$

## 3.2 Clustering with K-means

The attributes of this work is the clustering of vectors descriptors of the images to improve the classification by Bayesian Network by optimizing its construction.
For each object extracted of the image, we compute the two features describing a given object: color and textural features. The feature attributes are calculated for each image of the data-base and then clustered into k cluster.
We use the method of k-means to cluster the descriptor as shown in figure 6.

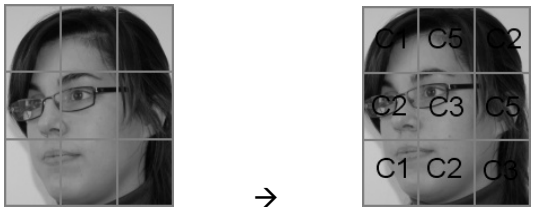

Fig. 6 The result of clustering with k-means into five clusters

## 3.3 Classifiers and learning

In this study, we utilize Naïve Bayes, Tree Augmented Naïve Bayes (TAN) and Forest Augmented Naïve Bayes (FAN) classifiers. This sub-section describes the implementation of these methods

### 3.3.1 Structure learning
- Naïve Bayes Network (NB)

A variant of Bayesian Network is called Naïve Bayes. It is one of the most effective and efficient classification algorithms. Figure 7 shows graphically the structure of naïve Bayes, each attribute node has the class node as it parent, but does not have any parent from attribute node. As the values of $P(a_i|c)$ can be easily calculated from training instances, naïve Bayes is easy to construct.

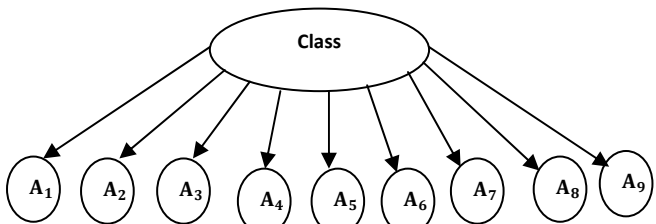

Fig.7 Naïve Bayes (NB)

- Tree Augmented Naïve Bayes (TAN)

It is obvious that the conditional independence assumption in naïve Bayes is in fact rarely true. Indeed, naive Bayes has been found to work poorly for regression problems (Frank et al., 2000), and produces poor probability estimation(Bennett, 2000). One way to alleviate the conditional independence assumption is to extend the structure of naive Bayes to explicitly represent attribute dependencies by adding arcs between attributes.
Tree augmented naive Bayes (TAN) is an extended tree-like naive Bayes (Friedman et al., 1997), in which the class node directly points to all attribute nodes and an attribute node can have only one parent from another attribute node (in addition to the class node). Figure 8 shows an example of TAN. In TAN, each node has at most two parents (one is the class node).

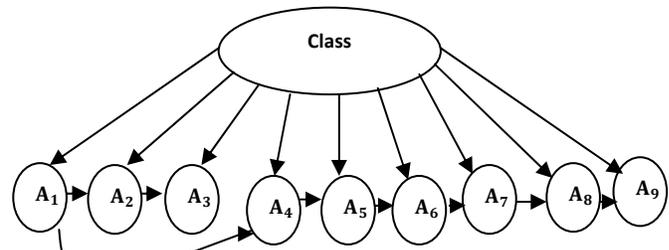

Fig. 8 Tree Augmented Naïve Bayes (TAN)

Construct-TAN procedure is described here as a blend of Friedman et al.'s and Perez et al.'s work. The procedure uses conditional mutual information, between two features given the class variable, which is described as follows:

$$I_p(A_i, A_j|C) = \sum_{\substack{x \in A_i \\ y \in A_j \\ z \in C}} P(x, y, z) \log \frac{P(x, y|z)}{P(x|z)P(y|z)}$$

$A_i$ and $A_j$ are sets of feature variables and $C$ is the set of class variable, and x, y and z are the values of variables X, Y and Z respectively. This function measures the information provided by $A_i$ and $A_j$ for each other when the class variable is given.

---

**Algorithm** : Construct-TAN

---

**Require:** Naïve Bayes, training set
**Ensure:** Structure of TAN
**1:** Calculate the conditional mutual information $I_p(A_i, A_j|C)$ between each pair of attributes, i≠j.
**2:** Build a complete undirected graph in which nodes are attributes $A_i, \dots, A_n$. Annotate the weight of an edge connecting $A_i$ to $A_j$ by $I_p(A_i, A_j|C)$.
**3:** Build a maximum weighted spanning tree.
**4:** Translate the resulting undirected tree to a directed one by choosing a root attribute and setting the direction of all edges to be outward from it.
**5:** Construct a TAN model by adding a vertex labeled by C and adding an arc from C to each$A_i$.

- **Forest Augmented Naïve Bayes (FAN)**

The experiments show that classification accuracy in TAN is higher than in Naïve Bayes. Two factors contribute to this fact. First, some edges may be unnecessary to exist in a TAN. The number of the edges is fixed to n-1. Sometimes, it might outfit the data. Second, in step 4, the root of the tree is randomly chosen and the directions of all edges are set outward from it. However, the selection of the root attribute is important because it defines the structure of the resulting TAN, since a TAN is a directed graph. It is interesting that the directions of edges in a TAN do not significantly alter modify the classification accuracy.

So, we correspondingly modify the TAN algorithm. First, we remove the edges that have conditional mutual information less than a threshold. To our understanding, these edges have a high risk to outfit the training data, and thus undermine the probability estimation. More precisely, we use the average conditional mutual information Iavg, defined in Eq.(1), as the threshold.

$$Iavg = \frac{\sum_i \sum_{j,j \neq i} I_p(A_i; A_j|C)}{n(n-1)} \qquad (1)$$

Where n is the number of attributes.
Second, we choose the attribute $A_{root}$ with the maximum mutual information with class, defined by Eq.(2), as the root:

$$A_{root} = argmax_{A_i} I_p(A_i; C) \qquad (2)$$

Where i=1,…,n. It is natural to use this strategy, since intuitively the attribute which has the greatest influence on classification should be the root of the tree

Since the structure of the resulting model is not a strict tree, we call our algorithm Forest augmented naive Bayes (FAN), described in detail as follows.

---

**Algorithm** : Construct-FAN

---

**Require:** TAN, training set
**Ensure:** Structure of FAN
**1:** Calculate the conditional mutual information $I_p(A_i, A_j|C)$ between each pair of attributes, i≠j, and calculate the average conditional mutual information $I_{avg}$,defined as follows:

$$Iavg = \frac{\sum_i \sum_{j,j \neq i} I_p(A_i; A_j|C)}{n(n-1)}$$

**2:** Build a complete undirected graph in which nodes are attributes $A_i, …, A_n$. Annotate the weight of an edge connecting $A_i$ to $A_j$ by $I_p(A_i, A_j|C)$.
**3:** Search a maximum weighted spanning tree.
**4:**Calculate the mutual information $I_p(A_i|C), i = 1,2,,…,n$ between each attribute and the class, and find the attribute

---

$A_{root}$ that has the maximum mutual information with class, as follows:

$$A_{root} = argmax_{A_i} I_p(A_i; C)$$

**5:** Transform the resulting undirected tree into a directed one by setting $A_{root}$ as the root and setting the direction of all edges to be outward from it.
**6:** Delete the directed edges with the weight of the conditional mutual information below the average conditional mutual information$Iavg$.
**7:** Construct a FAN model by adding a vertex labeled by C and adding a directed arc from C to each$A_i$.

### 3.3.2 Parameters learning

NB, TAN and FAN classifiers parameters were obtained by using the procedure as follows (Ben Amor 2006).
In the implementation of NB, TAN and FAN, we used the Laplace estimation to avoid the zero-frequency problem. More precisely, we estimated the probabilities $P(c), P(a_i|c)$ and $P(a_i|a_j, c)$ using Laplace estimation as follows.

$$P(c) = \frac{N(c) + 1}{N + k}$$
$$P(a_i|c) = \frac{N(c, a_i) + 1}{N(c) + v_i}$$
$$P(a_i|a_j, c) = \frac{N(c, a_i, a_j) + 1}{N(c, a_j) + v_i}$$

Where  -  N: is the total number of training instances.
- k: is the number of classes,
- $v_i$: is the number of values of attribute $A_i$,
- $N(c)$: is the number of instances in class c,
- $N(c, a_i)$: is the number of instances in class c and with $A_i = a_i$,
- $N(c, a_j)$: is the number of instances in class c and with $A_j = a_j$,
- $N(c, a_i, a_j)$: is the number of instances in class c and with $A_i = a_i$ and$A_j = a_j$.

### 3.3.3 Classification

In this work the decisions are inferred using Bayesian Networks. Class of an example is decided by calculating posterior probabilities of classes using Bayes rule. This is described for both classifiers.

- **NB classifier**

In NB classifier, class variable maximizing Eq.(3) is assigned to a given example.

$$P(C|A) = P(C)P(A|C) = P(C) \prod_{i=1}^{n} P(A_i|C) \qquad (3)$$

- TAN and FAN classifiers

In TAN and FAN classifiers, the class probability $P(C|A)$ is estimated by the following equation defined as:

$$P(C|A) = P(C) \prod_{i=1}^{n} P(A_i|A_j, C)$$

Where $A_j$ is the parent of $A_i$ and

$$\begin{cases} P(A_i|A_j, C) = \dfrac{N(c, a_i, a_j)}{N(c, a_j)} & \text{si } A_j \text{ existe} \\[2mm] P(A_i|A_j, C) = \dfrac{N(c, a_i)}{N(c)} & \text{si } A_j \text{ n'existe pas} \end{cases}$$

The classification criterion used is the most common maximum a posteriori (MAP) in Bayesian Classification problems. It is given by:

$$d(A) = \text{argmax}_{classe} P(classe|A)$$
$$= \text{argmax}_{classe} P(A|classe) \times P(classe)$$
$$= \text{argmax}_{classe} \prod_{i}^{N} P(A_i|classe) \times P(classe)$$

## 4. Experiments and Results

Now, we present the results of the contribution of our approach to classify images of some examples of classes from the database used 'Database of Faces'.

The ORL face database consists of ten different images of each of 40 distinct subjects with 92*112 pixels. For some subjects, the images were taken at different times, varying the lighting, facial expressions (open /closed eyes, smiling / not smiling) and facial details (glasses / no glasses) as shown at figure 9. All the images were taken against a dark homogeneous background with the subjects in an upright, frontal position (with tolerance for some side movement).

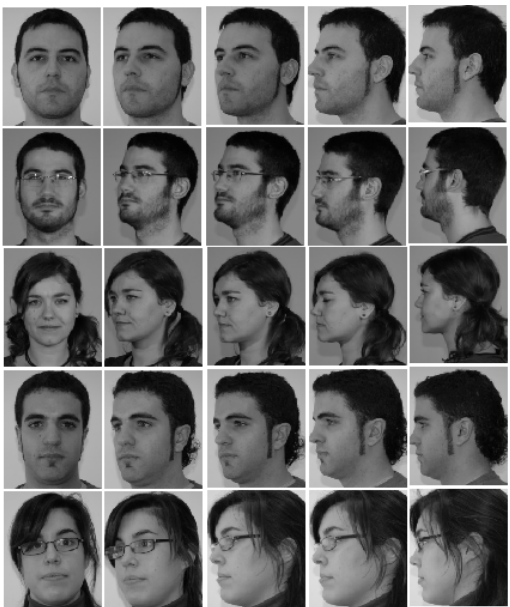

Fig. 9 Example 'Database of Faces'

## 4.1 Structure learning

We have used Matlab, and more exactly Bayes Net Toolbox of Murphy (Murphy, on 2004) and Structure Learning Package described in (Leray and al. 2004) to learn structure. Indeed, by applying the algorithm of TAN and FAN with different number of cluster, we obtained the structures as follow:

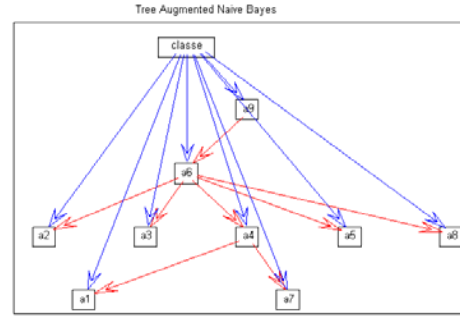

Fig. 10 Structure of TAN obtained with number of cluster k=5

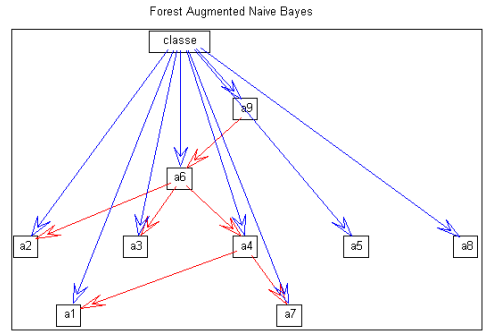

Fig. 11 Structure of FAN obtained with number of cluster k=5 and threshold S1

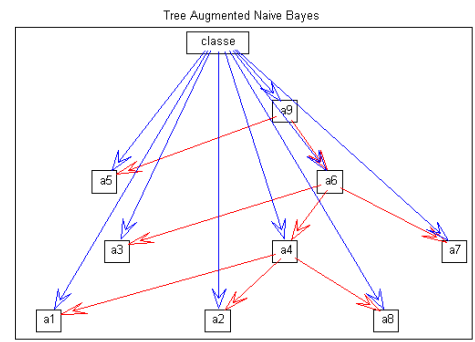

Fig.12 Structure of TAN obtained with number of cluster k=10

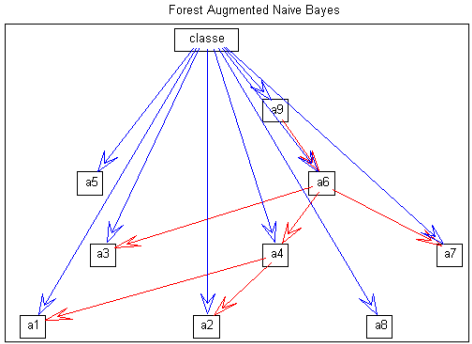

Fig. 13 Structure of FAN obtained with number of cluster k=10 and threshold S1

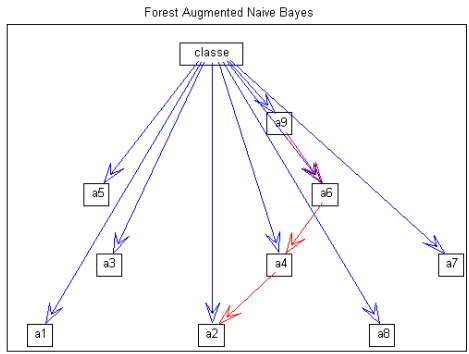

Fig.14 Structure of FAN obtained with number of cluster k=10 and threshold S2

## 4.2 Parameter learning

We estimated the conditional and a priori probability $P(c), P(a_i|c)$ and $P(a_i|a_j, c)$ using Laplace estimation. We have used the Laplace estimation described in section (3.3.2) to avoid the zero-frequency problem; we obtained the results as follows:

Table 1: a priori probability of attribute class P(class)

| P (class) | | | | |
|---|---|---|---|---|
| class 1 | class 2 | class 3 | class 4 | class 5 |
| 0,2 | 0,2 | 0,2 | 0,2 | 0,2 |

Table 2: a priori probability of $P(a_i|c)$ with number of cluster k=5

| P (a1\|c) | | | | | |
|---|---|---|---|---|---|
| | c1 | c2 | c3 | c4 | c5 |
| 1 | 0,80 | 0,84 | 0,28 | 0,84 | 0,76 |
| 2 | 0,04 | 0,04 | 0,04 | 0,04 | 0,04 |
| 3 | 0,04 | 0,04 | 0,04 | 0,04 | 0,04 |
| 4 | 0,08 | 0,04 | 0,60 | 0,04 | 0,12 |
| 5 | 0,04 | 0,04 | 0,04 | 0,04 | 0,04 |

## 4.3 Results

For each experiment, we used the percentage of correct classification (PCC) to evaluate the classification accuracy defined as:

$$PCC = \frac{\text{number of images correctly classified}}{\text{total number of images classified}}$$

Other criterion can be used, such as:

- P(X=C1 | class=C1) and P(X=C2 | class=C2) represent the rate of good classification.
- P(X=C1 | class=C2) and P(X=C2 | class=C1) represent the rate of bad classification.

The results of experiments are summarized in Table 3 and Table 4 with different number of clusters k=5 and k=10. Naive Bayes, TAN and FAN use the same training set and are exactly evaluated on the same test set.

Table 3: Classification accuracy of NB, TAN and FAN with number of cluster k=5

| | | class 1 | class 2 | class 3 | class 4 | class 5 | mean classification |
|---|---|---|---|---|---|---|---|
| NB | training set | 0,80 | 0,90 | 0,75 | 0,60 | 0,55 | 0,72 |
| | test set | 0,40 | 0,80 | 0,80 | 0,50 | 0,60 | 0,62 |
| TAN | training set | 0,15 | 0,90 | 0,40 | 0,05 | 0,10 | 0,31 |
| | test set | 0,10 | 0,70 | 0,50 | 0,10 | 0,10 | 0,30 |
| FAN | training set | 0,40 | 0,65 | 0,30 | 0,05 | 0,25 | 0,33 |
| | test set | 0,30 | 0,70 | 0,40 | 0,10 | 0,10 | 0,32 |
| FAN (new threshold) | training set | 0,75 | 0,95 | 0,85 | 0,70 | 0,60 | 0,77 |
| | test set | 0,30 | 0,80 | 0,80 | 0,70 | 0,60 | 0,64 |

As shown in experiments results in Table 3, the Naïve Bayes performed better than Tree Augmented Naïve Bayes (TAN) and Forest Augmented Naïve Bayes (FAN). However, when we increase the threshold in FAN we find out that the rate of mean classification is improved and became better than NB and TAN.

Table 4: Classification accuracy of NB, TAN and FAN with number of cluster k=10

| | | class 1 | class 2 | class 3 | class 4 | class 5 | mean classification |
|---|---|---|---|---|---|---|---|
| RN | training set | 0,95 | 1,00 | 0,80 | 0,95 | 0,85 | 0,91 |
| | test set | 0,40 | 1,00 | 0,50 | 0,60 | 0,80 | 0,66 |
| TAN | training set | 0,60 | 0,75 | 0,45 | 0,35 | 0,45 | 0,52 |
| | test set | 0,10 | 0,90 | 0,40 | 0,30 | 0,40 | 0,42 |
| FAN | training set | 0,75 | 0,90 | 0,25 | 0,55 | 0,40 | 0,57 |
| | test set | 0,30 | 0,90 | 0,40 | 0,60 | 0,30 | 0,50 |
| FAN (new threshold) | training set | 0,95 | 1,00 | 0,85 | 0,90 | 0,90 | 0,92 |
| | test set | 0,30 | 0,90 | 0,70 | 0,80 | 0,90 | 0,72 |

As shown in experiments results in Table 4, the classification accuracy with number of cluster equal to 10 performed better than classification accuracy with number of cluster equal to 5. So, we were obliged to study the effect of variation of number of cluster in the classification accuracy.

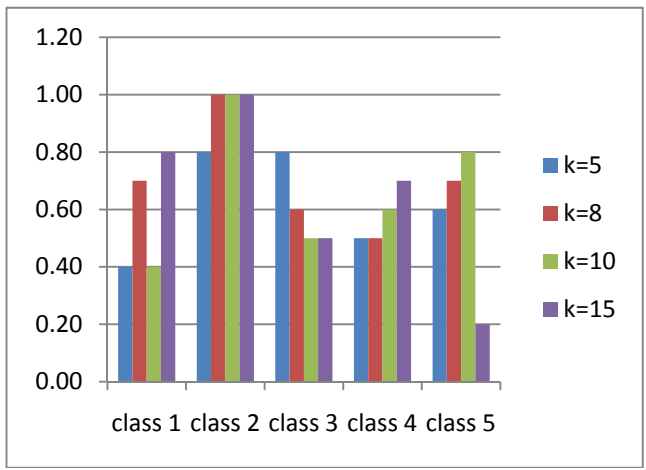

Fig.15 Variation of classification accuracy saved by Naïve Bayes with different number of cluster (k=5, 8, 10, and 15)

As experiments results shown in figure 15, we see that the mean of classification accuracy saved by Naïve Bayes is higher when number of cluster k=8 and then it decreased. So, we concluded that number of cluster is optimal when k=8 that means the number of cluster in this case reflect better the data.

## 5. Conclusion

In this study we have presented a new approach for classifying image of faces by using Bayesian Network and K-means to cluster the vectors descriptors of the images to improve the classification using Bayesian Network by optimizing its construction. We have implemented and compared three classifiers: NB, TAN and FAN. The goal was to be able to apply algorithms that can produce useful information from a high dimensional data. In particular, the aim is to improve Naïve Bayes by removing some of the unwarranted independence relations among features and hence we extend Naïve Bayes structure shown in figure 7 by implementing the Tree Augmented Naïve Bayes. Unfortunately, our experiments show that TAN performs even worse than Naïve Bayes in classification. Responding to this problem, we have modified the traditional TAN learning algorithm by implementing a new learning algorithm, called Forest Augmented Naïve Bayes. We experimentally test our algorithm in data image of faces and compared it to NB and TAN. The experimental results show that FAN improves significantly NB classifiers' performance in classification. In addition, the results show that the mean of classification accuracy is better when the number of cluster is optimal that means the number of cluster that can reflect better the data.

## 6. References

[1] A. Mojsilovic, "A computational model for color naming and describing color composition of images", IEEE Transactions on Image Progressing, 2005, Vol.14, No.5, pp.690-699.

[2] Q. Zhang and I. Ebroul, "A bayesian network approach to multi-feature based image retrieval", First International Conference on Semantic and Digital Media Technologies, GRECE, 2006.

[3] P.S. Rodrigues and A.A. Araujo, "A bayesian network model combining color, shape and texture information to improve content based image retrieval systems", 2004

[4] M.A. Mahjoub and K. Jayech, "Indexation de structures de documents par réseaux bayésiens", COnférence en Recherche d'Infomations et Applications, CORIA, 2010, Tunisia, pp.163-178

[5] S. Aksoy and R. Haralik, "Textural features for image database retrieval", IEEE Workshop on Content-Based Access of Image and Video Libraries, June, 1998, USA.

[6] C. Biernacki and R.Mohr, "Indexation et appariement d'images par modèle de mélange gaussien des couleurs ", Institut National de Recherche en Informatique et en Automatique, Rapport de recherche, No.3600, Janvier, 1999.

[7] E.Lebarbier and T.M.Huard, " Le critère BIC: fondements théoriques et interprétation ", Institut National de Recherche en Informatique et en Automatique, Rapport de recherche, No.5315, Septembre, 2004.

[8] A. Sharif and A. bakan, "Tree augmented naïve baysian classifier with feature selection for FRMI data", 2007.

[9] N.Ben Amor, S. Benferhat, Z. Elouedi, " Réseaux bayésiens naïfs et arbres de décision dans les systèmes de détection d'intrusions " RSTI-TSI, Vol. 25, No.2, 2006, pp.167-196.

[10] D.M.Chickering, "Learning bayesian network is NP-complete", In Learning from data: artificial intelligence and statistics, pp.121-130,New York, 1996.

[11] J.Cerqudes and R.L.Mantaras, "Tractable bayesian learning of tree augmented naïve bayes classifiers", January, 2003

[12] J.F.Cooper, "Computational complexity of probabilistic inference using bayesian belief networks", Artificial Intelligence, Vol.42, pp.393-405, 1990.

[13] N.Loménie and N.Viencent, R.Mullot, "Les relations spatiales : de la moélisation à la mise en œuvre", Revue des nouvelles technologies de l'information, cépadues-éditions ,2008

[14] N.Friedman and N.Goldszmidt, "Building classifiers using bayesian networks", Proceedings of the American association for artificial intelligence conference, 1996.

[15] O.Francois, De l'identification de structure de réseaux bayésiens à la reconnaissance de formes à partir d'informations complètes ou incomplètes, Thèse de doctorat, Institut National des Sciences Appliquées de Rouen, 2006.

[16] O.Francois and P.Leray, "Learning the tree augmented naïve bayes classifier from incomplete datasets", LITIS Lab., INSA de Rouen, 2008.

[17] F.Hsu , Y.Lee, S.Lin, "2D C-Tree Spatial Representation for Iconic Image", Journal of Visual Languages and Computing, pp 147-164, 1999.

[18] G.Huang, W.Zhang, L.Wenyin, "A Discriminative Representation for Symbolic Image Similarity Evaluation", Workshop on Graphics Recognition, Brazil, 2007.

[19] P. Leray, "Réseaux Bayésiens : apprentissage et modélisation de systèmes complexes", novembre, 2006.

[20]X.Li, " Augmented naive bayesian classifiers for mixed-mode data", December, 2003.

[21] P.Naim, P.H.Wuillemin, P.Leray, O.Pourret, A.Becker. Réseaux bayésiens, Eyrolles, Paris, 2007.

[22] E.G.M.Patrakis, "Design and evaluation of spatial similarity approaches for image retrieval", In Image and Vision Computing Vol.20, pp.59-76, 2001.

[23] L.Smail , "Algorithmique pour les réseaux Bayésiens et leurs extensions ", Thèse de doctorat, 2004

[24] J.E. Meng, W. Chen, W. Shiqian, "High speed face recognition based on discrete cosine transform and RBF neural networks"; IEEE Transactions on Neural Networks, Vol. 16, No.3, pp.679 − 691, 2005.

[25] M.Turk and A. Pentland. "Eigenfaces for recognition". Journal of Cognitive Neuroscience, Vol.3, pp.71-86, 1991.

[26] M.Jones and P.Viola, "Face Recognition using Boosted Local Features", IEEE ICCV,2003.

[27] A.S.Mian, M.Bennamoun, R.A.Owens, "2D and 3D Multimodal Hybrid Face Recognition", pp. 344–355, 2006.

[28] W.Zhao, R.Chellappa, P.J.Phillips, A.Rosenfeld, "Face Recognition: A Literature Survey", ACM Computing Survey, pp.399–458, 2003.

[29] J.Huang, B.Heisele, V.Blanz, "Component-based Face Recognition with 3D Morphable Models",2003.

[30] C.Garcia, G.Zikos, G.Tziritas, "A Wavelet-based Framework for Face Recognition",Workshop on advances in facial image analysis and recognition technology, 5th European conference on computer vision,1998.

[31] M. Turk and A. Pentland, "Face Recognition Using Eigenfaces", In IEEE Intl. Conf on Computer Vision and Pattern Recognition , pp. 586–591, 1991

[32] M. Bartlett, J. Movellan, T. Sejnowski. Face Recognition by Independent Component Analysis. IEEE Trans. on Neural Networks, Vol.13,No.6,pp.1450–1464, 2002.

[33] P. Belhumeur, J. Hespanha, D. Kriegman. Eigenfaces vs. Fisherfaces: Recognition Using Class Specific Linear Projection. IEEE Trans. on Pattern Analysis and Machine Intelligence, Vom.19, No.7, pp.711–720, 1997.

[34] T. Ahonen, A. Hadid, M. Pietikâinen, "Face Recognition With Local Binary Patterns", In European Conference on Computer Vision (ECCV), pp. 469–481, 2004.

[35] F. Cardinaux, C. Sanderson, S. Marcel, "Comparison of MLP and GMM classifiers for face verification on XM2VTS", In 4th Intl. Conf. Audio- and Video-based Biometric Person Authentication, AVBPA, Vol.2688, pp.1058-1059, 2003.

[36] S. Lucey and T. Chen, "A GMM Parts Based Face Representation for Improved Verification through Relevance Adaptation", In IEEE Intl. Conf on Computer Vision and Pattern Recognition (CVPR), pp. 855–861, 2004.

[37] Y. Rodriguez and S. Marcel, "Face Authentication Using Adapted Local Binary Pattern Histograms", In European Conference on Computer Vision (ECCV), pp. 321–332, 2006.

[38] L. Wiskott, J.M. Fellous, N. Kruger, C. Von Der Malsburg, "Face Recognition By Elastic Bunch Graph Matching", In Intelligent Biometric Techniques in Fingerprint and Face Recognition, pp. 355–396, 1999.

[39] A. Nefian and M. Hayes, "Hidden Markov Models for Face Recognition", In IEEE Intl. Conf. on Acoustics, Speech, and Signal Processing (ICASSP), Vol.5, pp.2721–2724, 1998.

[40] F.Cardinaux, C.Sanderson, S. Bengio, "User Authentication via Adapted Statistical Models of Face Images", IEEE Trans. on Signal Processing, Vol.54,pp.361– 373, 2005

[41] R.Huang, V.Pavlovic, D.N.Metaxas, "A Hybrid Face Recognition Method using Markov Random Fields" , Proceedings of the Pattern Recognition, 17th International Conference on (ICPR'04), Vol.3,2004.

[42] S.Sharavanan and M.Azath, "LDA based face recognition by using hidden Markov Model in current trends", International Journal of Engineering and Technology, Vol.1, pp.77-85, 2009.

**Khlifia Jayech** is a Ph.D. student in the department of Computer Science of National Engineering School of Sousse. She obtained her master degree from Higher Institute of Applied Sciences and Technology of Sousse (ISSATS), in 2010. Her areas of research include Data Retrieval, Bayesian Network, Hidden Markov Model and Recurrent Neural Network, and Handwriting Recognition.

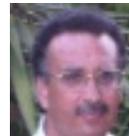

**Dr Mohamed Ali Mahjoub** is an assistant professor in the department of computer science at the Preparatory Institute of Engineering of Monastir. His research interests lie in the areas of HMM, Bayesian Network, Pattern Recognition, and Data Retrieval. His main results have been published in international journals and conferences.